%% file: main_CVPR25.tex
\documentclass[10pt,twocolumn,letterpaper]{article}
\usepackage[accsupp]{axessibility}  % Improves PDF readability for those with disabilities.

\usepackage[pagenumbers]{cvpr} % arXiv version

\input{preamble}

\makeatletter
\def\@fnsymbol#1{\ensuremath{\ifcase#1\or \dagger\or \ddagger\or
\mathsection\or \mathparagraph\or \|\or **\or \dagger\dagger
\or \ddagger\ddagger \else\@ctrerr\fi}}
\makeatother

\def\paperID{2233}
\def\confName{CVPR}
\def\confYear{2025}

\title{\ApproachName: Bootstrapped Object Placement with Detection Transformers}

\author{
Hang Zhou\textsuperscript{\rm 1} \qquad
Xinxin Zuo\textsuperscript{\rm 2} \qquad
Rui Ma\textsuperscript{\rm 3}\thanks{Corresponding author.} \qquad
Li Cheng\textsuperscript{\rm 1}\\
\textsuperscript{\rm 1}University of Alberta\quad \
\textsuperscript{\rm 2}Concordia University\quad \ 
\textsuperscript{\rm 3}Jilin University \
}

\begin{document}
\maketitle
\input{sec/0_abstract}
\input{sec/1_intro}
\input{sec/2_related}

\input{sec/3_method}
\input{sec/4_results}
\input{sec/5_conclusions}

\input{sec/6_acks}
{
    \small
    \bibliographystyle{ieeenat_fullname}
    \bibliography{main}
}
\input{sec/7_supp}

\end{document}

%% file: preamble.tex
\definecolor{cvprblue}{rgb}{0.21,0.49,0.74}
\usepackage[pagebackref,breaklinks,colorlinks,allcolors=cvprblue]{hyperref}

\usepackage{bbm}
\usepackage{orcidlink}
\usepackage{tikz}
\usepackage{wrapfig}
\newcommand*\circled[1]{\tikz[baseline=(char.base)]{
            \node[shape=circle,draw,inner sep=0.7pt] (char) {#1};}}
\DeclareMathOperator*{\argmax}{arg\,max}

\usepackage{color}
\definecolor{turquoise}{cmyk}{0.65,0,0.1,0.3}
\definecolor{purple}{rgb}{0.65,0,0.65}
\definecolor{dark_purple}{rgb}{0.5,0,0.5}
\definecolor{dark_green}{rgb}{0, 0.5, 0}
\definecolor{orange}{rgb}{0.8, 0.6, 0.2}
\definecolor{red}{rgb}{0.8, 0.2, 0.2}
\definecolor{darkred}{rgb}{0.6, 0.1, 0.05}
\definecolor{blueish}{rgb}{0.0, 0.3, .6}
\definecolor{light_gray}{rgb}{0.7, 0.7, .7}
\definecolor{pink}{rgb}{1, 0, 1}
\definecolor{greyblue}{rgb}{0.25, 0.25, 1}

\newcommand{\hang}[1]{{\color{black}{#1}}}

\newcommand{\ApproachName}{\textsc{BootPlace}\xspace}
\newcommand{\SupplementaryMaterial}{{{\color{dark_purple}supplementary}}\xspace}

%% file: sec/0_abstract.tex
\begin{abstract}

In this paper, we tackle the copy-paste image-to-image composition problem with a focus on object placement learning. 
Prior methods have leveraged generative models to reduce the reliance for dense supervision. However, this often limits their capacity to model complex data distributions.
Alternatively, transformer networks with a sparse contrastive loss have been explored, but their over-relaxed regularization often leads to imprecise object placement. 
We introduce \ApproachName, a novel paradigm that formulates object placement as a \textnormal{placement-by-detection} problem. 
Our approach begins by identifying suitable regions of interest for object placement. This is achieved by training a specialized detection transformer on object-subtracted backgrounds, enhanced with multi-object supervisions. 
It then semantically associates each target compositing object with detected regions based on their complementary characteristics. 
Through a boostrapped training approach applied to randomly object-subtracted images, our model enforces meaningful placements through extensive paired data augmentation. 
Experimental results on established benchmarks demonstrate \ApproachName's superior performance in object repositioning, markedly surpassing state-of-the-art baselines on Cityscapes and OPA datasets with notable improvements in IOU scores. 
Additional ablation studies further showcase the compositionality and generalizability of our approach, supported by user study evaluations. 
\hang{Code is available at \url{https://github.com/RyanHangZhou/BOOTPLACE}}

\end{abstract}

%% file: sec/1_intro.tex
\section{Introduction}
\label{sec:intro}

Compositional modeling is an emerging research area in both computer vision and computer graphics, focusing on constructing visual scenes by \textit{assembling} components, objects, or elements with precise \textit{placement} and interaction. 
Task-agnostic approaches, such as image-to-image~\cite{lin2018st,zhang2020learning}, 3D-object-to-image~\cite{hold2019deep,wang2022neural}, and 3D-object-to-3D-object~\cite{wei2023lego,para2023cofs} composition, have gained wide-ranging applications in content creation~\cite{zhu2023topnet} and data augmentation~\cite{antoniou2017data,dwibedi2017cut,shorten2019survey}.

The evolution of image-to-image composition as a standardized input-output interface has enabled precise image editing and enhanced downstream tasks such as semantic segmentation and object detection~\cite{ghiasi2021simple,zhou2022sac}.
Another promising direction involves image-guided composition using diffusion models~\cite{song2023objectstitch,ho2020denoising,lu2023tf,chen2024anydoor}, which excels in blending performance but often struggles to preserve the appearance and orientation of the composed objects.

\input{figs/teaser.tex}

A key challenge in precise placement learning lies in the difficulty of training models to accurately place objects due to \textit{sparse supervision} of ground-truth location labels. 
Early approaches addressed this by employing generative models~\cite{lin2018st,azadi2020compositional,zhang2020learning,zhou2022sac} conditioned on joint embeddings of scene images and object patches. These models generated placements using 2D spatial transformations of the object or its bounding box coordinates, bypassing the need for explicit and dense supervision through compositionality-aware saliency maps. 
Alternatively, transformer-based approaches~\cite{zhu2023topnet} have been used to regress bounding box placements by directly learning an object-to-background association. This introduces relaxed constraints to mitigate the scarcity of labels. 
A more recent study~\cite{SiyuanZhou2022LearningOP} increases label availability by manually annotating positive and negative composite pairs,  
enabling the network to learn to avoid suboptimal object placements. 
However, the reliance on human-annotation is labor-intensive and prone to inaccuracies, limiting scalability for large-scale applications.

In this paper, we present \textit{\ApproachName}, a versatile framework designed to \textit{detect} optimal object placement guided by object queries, and \textit{associate} target compositing object with their best-matched detected region. 
The detection module is trained on object-subtracted images to identify regions of interest. 
Region codes learned from this module are independently decoded into bounding-box coordinates and class labels via a feedforward network, resulting in multiple placement predictions. 
Moreover, to associate the compositing object with its best-matched detected region, we develop a dedicated association network guided by localized semantic complementarity. 
To address the challenge of label sparsity, we propose a boostrapped training strategy using randomly object-subtracted images, which enhances data diversification and enables precise object placement.

Our contributions are summarized as follows:
\begin{description}
\item[Placement-by-detection framework:] We introduce a novel approach for precise object placement by employing detection constraints to identify regions of interest.
\item[Object-to-region association loss:] We establish visual correspondences between object queries and local regions of interests in object-subtracted images through an object-to-region associating loss mechanism.
\item[Multi-object placement pipeline:] We implement a multi-object placement pipeline, supervised by multi-labels of intact objects, as well as a bootstrapped training strategy to enhance data diversification and improve placement accuracy. 
\end{description}

%% file: figs/teaser.tex
\begin{figure}[t!]
\vspace{-2.5mm}
\begin{center}
\includegraphics[width=1.0\linewidth]{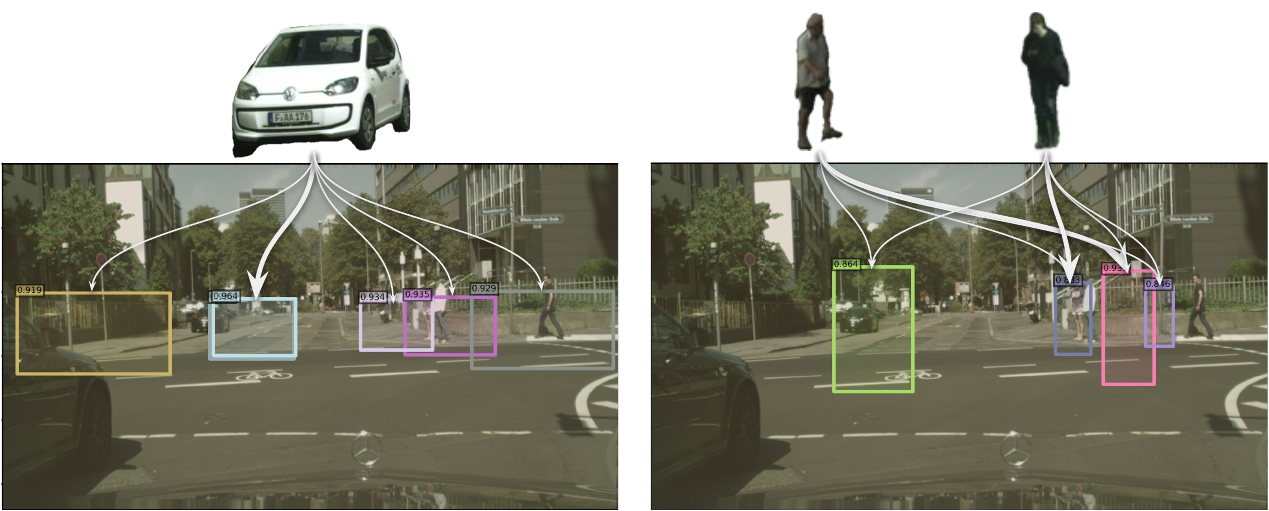}
\end{center}
\vspace{-2.5mm}
\caption{
Our approach, \ApproachName{}, \textit{detects} regions of interest (represented as bounding boxes) for object composition and \textit{assigns} each target object to its best-matched detected region. 
Each object is connected to each detected region with weighted connections, with the bold arrow indicating the strongest link. 
}
\label{fig:teaser}
\vspace{-4.0mm}
\end{figure}

%% file: sec/2_related.tex
\section{Related works}

\paragraph{Image composition.}
Apart from classic methods, including alpha matting~\cite{smith1996blue} and Poisson image editing~\cite{perez2023poisson}, which seamlessly blend two images, recent works~\cite{tsai2017deep,cong2020dovenet} have focused on \textit{harmonizing} the composited region by adjusting its color to better match the background. 
More recently, there has been a trend towards guided inpainting with visual prompts~\cite{meng2021sdedit,song2023objectstitch,chen2024anydoor} as a means to achieve more realistic composites. 
Despite the capabilities of various models to merge two images naturally, these methods all require a bounding box to indicate the target location for the paste. 
\hang{Canet et al.~\cite{canet2025thinking} proposed unconstrained compositing by jointly learning object placement and composition.}
In contrast, our model aims to generate an optimal placement \hang{applicable to}
a variety of composition models.

\paragraph{Object placement.}
The recent literature has seen significant advancement in the field of learning object placement for copy-paste object compositing, as evidenced by multiple notable works~\cite{lin2018st,chen2019toward,li2019putting,zhan2019spatial,zhan2019spatial,azadi2020compositional,zhang2020learning,zhou2022sac,SiyuanZhou2022LearningOP,zhu2023topnet,liu2024diffpop,niu2022fast}. 
ST-GAN~\cite{lin2018st} employed an iterative spatial transformation network to generate a 2D transformation for object placement via adversarial training against a natural image manifold. 
GCC-GAN~\cite{chen2019toward} incorporated geometric and color consistency in its approach for object placement. 
Li et al.~\cite{li2019putting} represented human poses and placement as affordance modeled by variational autoencoders (VAEs). 
SF-GAN~\cite{zhan2019spatial} fused geometric and appearance realism within its generator network. 
Compositional-GAN~\cite{azadi2020compositional} developed a self-consistent composition-by-decomposition network with a viewpoint-aware appearance flow network to enhance placement. 
PlaceNet~\cite{zhang2020learning} introduced spatial diversity loss for multi-object placement to achieve greater image realism. 
GracoNet~\cite{SiyuanZhou2022LearningOP} treated object placement as a graph completion problem and employed adversarial training with annotated positive and negative composite pairs. 
SAC-GAN~\cite{zhou2022sac} proposed conditional VAE-GAN for structural coherence via adversarial training against the semantic manifold. 
TopNet~\cite{zhu2023topnet} designed a transformer~\cite{vaswani2017attention} network for object-background correlation with a sparse contrastive loss on placement. 
DiffPop~\cite{liu2024diffpop} proposed a plausibility-guided denoising diffusion model to learn the scale and spatial relations among multiple objects and the corresponding scene image. 
Copy-paste composition also advocates data augmentation for enhancing downstream tasks such as object detection and semantic segmentation~\cite{ghiasi2021simple,tripathi2019learning,zhang2020learning,zhou2022sac}. 
In contrast, our approach learns to detect potential locations for object placement and then associates the compositing object with the best-matched region, demonstrating superior ability to model complex scenes.

\paragraph{Object detection.}
Modern object detection can be broadly classified into two: two-stage~\cite{cai2018cascade,girshick2015fast,girshick2014rich,lin2017feature,ren2015faster} \textit{vs} single-stage detectors~\cite{liu2016ssd,redmon2016you,tian2019fcos}. 
\hang{Typically, these detectors involve heuristic steps such as proposal detection, thresholding, and non-maximum suppression.}
A significant progress was introduced by DETRs~\cite{carion2020end,kamath2021mdetr,li2022grounded,wang2022omni,zhang2022dino,bouniot2023proposal}, which explored end-to-end object detection using transformers. They treated detection as a set-to-set prediction problem and removed the need for heuristics. 
In our method, we propose a placement-by-detection paradigm and leverage the detection transformer to enhance the detection ability of the regions of interest.

%% file: sec/3_method.tex
\section{Approach}

\input{figs/inference.tex}

Our \textit{placement-by-detection} method achieves precise object placement for compositing objects of arbitrary classes. 
As shown in \Cref{fig:inference}, our approach is structured in two modules: (1) Learning a regions-of-interest detection network for identifying and localizing keyzones suitable for object placement (see \Cref{section:3.1}), and (2) Leveraging the detected keyzones, learning an object-to-region associating network (see \Cref{section:3.2}).

\subsection{Regions-of-interest detection}
\label{section:3.1}

Let $I$ be an image. 
The goal of regions-of-interest detection is to identify and localize attentive keyzones. 
Object detection is its special case where each region of interest is occupied by an individual object. 
A detector~\cite{carion2020end,ren2015faster,tian2019fcos} takes the image $I$ as input and produces $N$ region of interest $\{p_i\}$ with locations $\{b_i\}$, $b_i \in \mathbb{R}^4$ and a classification score $s_i \in \mathbb{R}^C$ from a set of predefined classes $C$ as its output. 
In the first stage, \ApproachName detects a fixed set of regions of interest from an object-subtracted image~$I$. 
Acting as the \textit{occluded context} behind objects, these regions are the \textit{inpainted} version by \textit{removing} objects from the source image. 
To initiate the object-centric image decomposition process, we perform a series of automated post-processing steps to decompose the source image into an object-subtracted image and a collection of intact object patches: 
1) Identify objects with instance segmentation models. 2) Remove objects using dilated masks with image inpainting models. 3) Smooth the image with Gaussian filter to eliminate inpainting artifacts.
This image decomposition process is visualized in \Cref{fig:network} and detailed in \SupplementaryMaterial.

Our detection model is built upon detection transformers, consisting of a CNN backbone for generating a lower-resolution image embedding, a multi-layer Transformer encoder to produce a compact image embedding, a multi-layer Transformer decoder that applies multi-headed self- and encoder-decoder attention using a set of learnable queries to generate output embeddings, and multiple prediction heads. 
To prevent compositing objects from being placed over existing scene objects, the locations of these scene objects are encoded into the network to guide placement. 
Specifically, scene object locations are encoded by an MLP-based location encoder, then concatenated with image features to form location-aware image features.

\input{figs/network.tex}

\subsection{Object-to-region associating network}
\label{section:3.2}

In addition to the detection objective, we aim to equip our detector with the ability to understand the relations between regions of interest and compositing objects. 
In the second stage, \ApproachName learns an object-to-region associating network using the detected keyzones and corresponding object patches from the first stage.

Our associating network associates regions of interest with object patches in a probabilistic and differentiable manner. 
Formally, each object query $q_k$ links region of interest $p_i$, and produces an object association score vector $g \in \mathbb{R}^N$ over a collection of region of interest. 
This association score vector then yields an association $\alpha_k \in \{\emptyset,1,2,...,N\}$ where $N$ is the number of detected regions of interest in the image $I$. 
The association then links ground-truth location $\tau_k$ to current detected regions of interest $p_i$, and is given by:
\begin{equation}
\tau_k=\left\{\begin{array}{ll}
\emptyset, & \text { if } \alpha=\emptyset \\
b_{\alpha_k}, & \text { otherwise }
\end{array}\right.
\end{equation}
where $\alpha=\emptyset$ indicates no association. 
Notably, the association step is differentiable and can be jointly trained with the underlying detector network.

The association score is elaborated as follows. 
Let $p_1, ..., p_N$ be a set of high-confidence regions of interest for image $I$. 
Let $B = \{b_1, ..., b_N\}$ be their corresponding
bounding boxes. 
Let $f_i \in \mathbb{R}^D$ be the $D$-dimensional features extracted from each region of interest. 
For convenience let $F=\{f_1, ..., f_N\}$ be the set of all detection features of image $I$. 
The associating network takes object features $F$ and an image $I$, and produces an object-specific association score $g(q_k,F) \in \mathbb{R}^N$. 
Let $g_i(q_k,F) \in \mathbb{R}$ be the score of the $i$-th region of interest in the image. A special output token $g_{\emptyset}(q_k,F)=0$ indicates no association. The associating network then outputs a distribution of associations over all regions of interest in the image for each object patch $q_k$. We model this by softmax activation: 
\begin{equation}
P_A(\alpha=i|F)=\frac{\exp{(g_i(q_k,F))}}{\sum_{j\in\{\emptyset, 1,2,...,N\}}\exp{(g_i(q_k,F))}}.
\end{equation}
Instead of just a dot-product association score between the object query and region-of-interest feature $g_i(q_k,F)$, to avoid the association with scene objects, we propose incorporating negative correlations rather than positive correlations for \textit{semantic complementary}, providing the model with a nuanced understanding and enhancing its ability to recognize optimal placement regions within the scene.
\begin{equation}
g_i(q_k,F)=-q_k\cdot F_i/\mu,
\end{equation}
where $\mu$ is a temperature parameter.
During training, we maximize the $\log$-likelihood of the ground-truth object-to-region association. During inference, we use the likelihood to find the best-matched object-to-region association.

\paragraph{Network structure.}
The object-to-region association comprises a CNN-based object encoder network for extracting object embeddings, and an association module that takes a stack of region-of-interest features $F \in \mathbb{R}^{N\times D}$ and a matrix of object query features $Q \in \mathbb{R}^{T\times D}$ as two inputs and produces an association matrix $G \in \mathbb{R}^{T\times N}$ between queries and regions of interest with cosine similarity.

\subsection{Training}
Given ground-truth bounding box locations $\tau_1, \tau_2, ..., \tau_T$ where $\tau_k\in \mathbb{R}^4\cup \emptyset$ and corresponding classes $c_1, c_2, ..., c_T$, the goal is to learn an associating network that estimates $P_A$. 
The associating network and the detection transformer are jointly trained, 
with the association network functioning as a region-of-interest head, similar to two-stage detectors~\cite{ren2015faster}. During each training iteration, high-confidence regions of interest
$b_1, ..., b_N$ along with their corresponding features $F$ and class prediction $s_1, s_2, ..., s_N$ are first obtained. 
For each sample, $P_A(\alpha|F)$ is then maximized.

Following DETR~\cite{carion2020end}, both the class prediction and the similarity of predicted and ground-truth boxes are taken into account for the assignment rule. 
Specifically, each element $k$ of the ground-truth set can be seen as a $(c_k,\tau_k)$ where $c_k$ is the target class label (which may be $\emptyset$). 
For the prediction with index (association) $\alpha_k$, the probability of class $c_k$ is defined as $P_{\alpha_k}(c_k)$ and
the predicted box as $b_{\alpha_k}$.

With these notations, the association cost between the ground truth and the prediction with index $\alpha_k$ is defined as
\begin{align}
L_{cost}(\tau_k, c_k, P_{\alpha_k}(c_k), b_{\alpha_k}) = & -\mathbbm{1}_{c_k\neq \emptyset}P_{\alpha_k}(c_k) \notag \\
& + \mathbbm{1}_{c_k\neq \emptyset}L_{box}(\tau_k, b_{\alpha_k}).
\end{align}

To find a one-to-one bipartite matching between predicted boxes and ground-truth boxes, a permutation of $T$ elements with the lowest cost is searched:
\begin{equation}\label{eq:1}
\hat{\alpha} = \argmax_{\alpha}\sum_k^T L_{cost}(\tau_k, c_k, P_{\alpha_k}(c_k), b_{\alpha_k}).
\end{equation}
This optimal assignment $\hat{\alpha}$ could be computed efficiently with the Hungarian algorithm, and is used to both train the bounding box regression of the underlying two-stage detector, and our assignment likelihood $P_A$.

\paragraph{Boostrapped training strategy.}
To enhance the diversity of training data, a bootstrapped strategy is employed for augmenting training data, as shown in \Cref{fig:network}. 
Specifically, having $T$ intact objects subtracted from an image, a subset of objects is randomly selected and is recomposed back to the object-subtracted image to create a randomly-object-subtracted image $\mathcal{I}$. 
The remaining objects are viewed as target compositing objects. 
Through this process, we expand the training data from a single scenario to a comprehensive set of combinations, reaching a total of $\sum_{i=1}^{T}\tbinom{T}{i}$ samples for each scene. 
This strategy significantly enriches the training dataset, exposing the model to a more extensive variety of scenarios.

The overall training losses includes the assignment from \Cref{eq:1} and an association loss for all compositing objects. 
The association loss is defined as the sum of the log-likelihood of the assignments $\alpha_k$ of each location $\tau_k$:
\begin{equation}
L_{asso}(F,\{\tau_1,\tau_2,...,\tau_T\})=-\sum_{k=1}^T\log P_A(\hat{\alpha}_k|F).
\end{equation}

We train $\mathcal{L}_{asso}$ jointly with standard detection losses~\cite{carion2020end}, including classification loss $\mathcal{L}_{cls}$ and bounding-box regression loss $\mathcal{L}_{box}$~\cite{rezatofighi2019generalized}: 
\begin{equation}
\mathcal{L}=\mathcal{L}_{cls}+\alpha\mathcal{L}_{box}+\beta\mathcal{L}_{asso}.
\end{equation}

%% file: figs/inference.tex
\begin{figure*}[t!]
\vspace{-2.5mm}
\begin{center}
\includegraphics[width=1.0\linewidth]{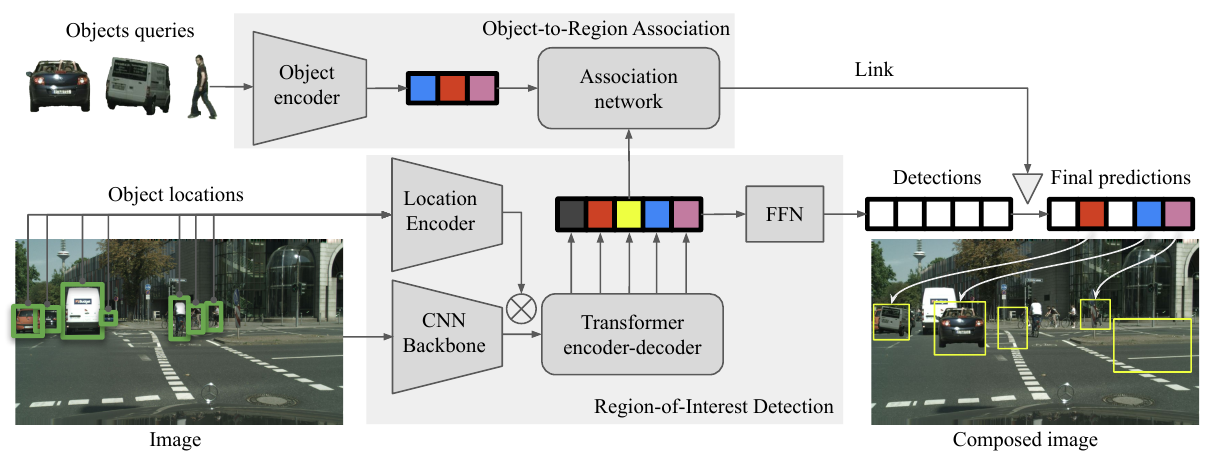}
\end{center}
\vspace{-6.0mm}
\caption{
\textbf{Network inference.}
Given a target image, several object queries (e.g., two cars and a pedestrian) and scene object locations, \ApproachName detects a set of candidate region of interest and associates each object with the best-fitting region, which are used to produce the composite image. 
\hang{$\otimes$ is feature concatenation and $\bigtriangledown$ is region-wise product. }
}
\label{fig:inference}
\vspace{-3.5mm}
\end{figure*}

%% file: figs/network.tex
\begin{figure*}[t!]
\vspace{-2.5mm}
\begin{center}
\includegraphics[width=1.0\linewidth]{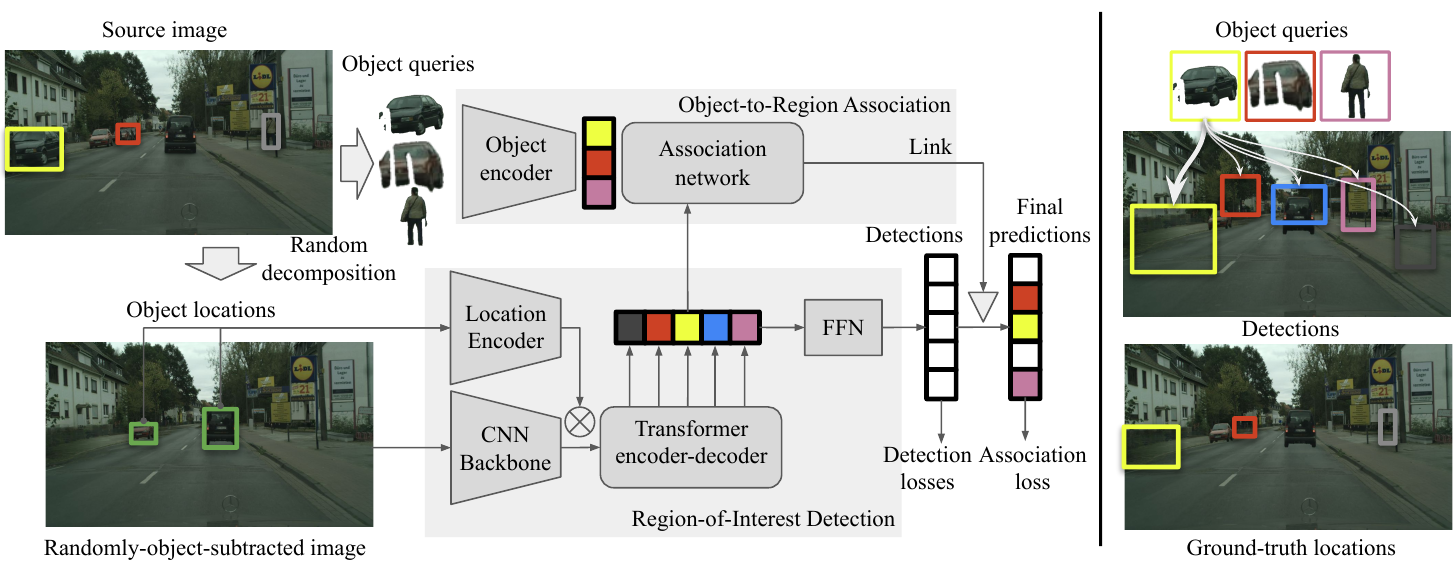}
\end{center}
\vspace{-6.0mm}
\caption{
\textbf{Network architecture and training.}
We prepare training data by first decomposing a source image into a randomly-object-subtracted image $\mathcal{I}$ and a set of object queries. 
During training, image $I$ and scene object locations are both fed into a detection transformer for region-of-interest detection. 
The object queries are fed into an association network for object-to-region matching, where the generated association links each object query with the detected region of interest. Losses comprises of detection loss and association loss.  
At the high level, we visualize the relations among object queries, detected regions of interest and ground-truth locations on the right side, where the best-matched association arrow is highlighted in bold. 
}
\label{fig:network}
\vspace{-1.5mm}
\end{figure*}

%% file: sec/4_results.tex
\section{Experiments}
\label{sec:expe}

\paragraph{Datasets.} 
We conducted experiments on Cityscapes~\cite{cordts2016cityscapes} and OPA~\cite{liu2021OPA} datasets. Cityscapes is a large-scale urban street dataset for semantic segmentation. 
We use MaskFormer~\cite{cheng2021per} to obtain individual objects with panoptic segmentation, and background images by removing objects and their shadows with pretrained LaMa inpainting model~\cite{suvorov2022resolution} and online PhotoKit~\footnote{https://photokit.com/} tool. 
After data cleaning, we build a multiple-object placement dataset containing 2,953 training images with 22,270 objects and their ground-truth labels, and 372 testing images with 2,713 objects. 
OPA is a human-annotated composite dataset collected from COCO~\cite{lin2014microsoft} dataset for object placement. It contains 62,074 training images and 11,396 test images without overlap. 
We only use positive samples, which consist of 21,350 pairs of images and objects for training and 3,566 pairs for testing.

\paragraph{Implementation details.}
We train \ApproachName with AdamW~\cite{loshchilov2018decoupled} optimizer with an initial learning rate of 0.0004, the backbones's to 0.00005 and weight decay to 0.0001. By default, we set $\alpha=5$, $\beta=1$, $\mu=0.07$, and Gaussian filter with $\sigma=5$.
Training on Cityscapes dataset takes 12 hours on one NVIDIA TITAN RTX GPU, and 8 hours for OPA dataset.

\paragraph{Evaluation.} 
We quantitatively evaluate object reposition accuracy using top-$k$ intersection-over-union (IOU) on all output bounding boxes and evaluate detection precision with IOU50. 
For object placement, 
\hang{we measure \textit{plausibility} by a user study involving 20 participants who compare the realism of composite images generated by different methods using the same object patches and backgrounds, and \textit{diversity} via standard deviation of bbox scale, x-center, and y-center with the trial number set to 5.}
Users were asked to focus on structural consistency instead of factors such as image resolution or color consistency. 
Unlike previous methods that relied on FID~\cite{heusel2017gans} or LPIPS~\cite{zhang2018unreasonable}, we find these metrics less accurate due to their inability to reflect regional boundary artifacts and resolution variations.

\input{tabs/recomposition.tex}
\input{figs/recomposition.tex}
\input{figs/recomposition_OPA.tex}
\input{figs/composition.tex}
\input{figs/generalizability.tex}

\subsection{Comparisons with previous methods}

\paragraph{Object reposition.}
Object reposition refers to object placement where the object patch and background are from the same image. 
In \Cref{table:recomposition}, we compare the object reposition accuracy with PlaceNet~\cite{zhang2020learning}, GracoNet~\cite{SiyuanZhou2022LearningOP}, SAC-GAN~\cite{zhou2022sac} and TopNet~\cite{zhu2023topnet}. We observe that \ApproachName shows significant performance advantages compared to other methods, with a 4\% improvement in top-5 IOU over TopNet, the state-of-the-art method. 
Our \ApproachName is more accurate at repositioning objects back onto backgrounds, in that it leverages the detection transformer's strong capability for precise localization. 
The visual results in \Cref{fig:recomposition,fig:recomposition_OPA} also illustrate the enhanced object placement performance achieved by our method. 
We notice that placing objects with distinct orientations, such as a car in a road scene, is more challenging than positioning generic objects as the orientation of the object must align with the semantic cues in the image, such as the direction of the road. 
It is also worth noting that when training our method on the OPA dataset, we only used single-object labels for supervision due to the lack of multi-object annotations. Despite this limitation, our method still performs better than others.

\input{tabs/composition.tex}
\input{figs/attention.tex}

\paragraph{Object placement.}
We perform object placement experiments on composing moving objects (e.g., cars, buses and trucks) into images from Cityscapes. 
In \Cref{table:composition}, we compare qualitative results with PlaceNet, SAC-GAN and TopNet, observing that
our \ApproachName significantly outperforms others. 
As shown in \Cref{fig:composition}, our method produces diverse and accurate placements, while previous methods often result in object collisions, incorrect placements, or inconsistent scales. 
GracoNet was excluded from evaluation as it requires negative composite images for training, which is unavailable in the Cityscapes dataset. 
For PlaceNet, we have dropped its diverse loss during training, as it is specifically designed for multi-object placement. 
In addition, we compare the \textit{scalability} of different models by testing the placement performance on Mapillary Vistas~\cite{neuhold2017mapillary} images using models trained on the Cityscapes dataset; see \Cref{table:composition} and \Cref{fig:generalizability}. 
Our \ApproachName exhibits superior scalability than other methods.

% removed a figure for top-k bbox detections (duplicate to another figure)
% \input{figs/overfit.tex}
% To assess this, we evaluate the IOU between the top-$k$ predicted bounding boxes and the ground-truth locations using a threshold of 0.1. 
% Based on 400 images from Cityscapes, we find that only 6.3\% of the predicted boxes closely match the ground-truth, suggesting our model does not overfit to inpainting artifacts. 
% The results are visualized in~\Cref{fig:overfit}, where solid boxes represent the predictions, which do not overlap with the dashed ground-truth boxes. For clarity, different colors are used to distinguish between objects.

\subsection{Overfitting to inpainting artifact?}
\input{figs/overfit.tex}
Although we have applied Gaussian blur and \hang{multi-region inpainting} to reduce the impact of inpainting artifacts on object placement, our model potentially overfits to the regions. 
In Figure~\ref{fig:reb_overfit}, we show the \textit{overfitting rate} across IOU thresholds for different top-$k$ predictions. 
At IOU50, the rate remains below 0.05, confirming our method exhibits minimal overfitting to artifacts.

\subsection{Ablation study}
\input{tabs/ablation.tex}
In \Cref{table:ablation}, we compare our method with several variations: a) no Gaussian smoothing; b) no bootstrapped data augmentation; c) using positive correlations; \hang{
d) single-object supervision; (e) no location encoder; (f) full model. }
The results support our design choices for individual modules. 
Gaussian smoothing is crucial in preventing overfitting to inpainting artifacts and pixel variations. 
Data augmentation leads to a \hang{0.17} improvement in \hang{IOU@5}. 
Negative correlations help our network avoid placing compositing objects over existing objects in the scene, thereby preventing unwanted occlusions. 
\hang{The full multiple-object supervision outperforms the single-object supervision, with IOU@5 increased by 0.042. 
The explicit location input significantly improves reposition accuracy by constraining the detection space, with IOU@5 increased by 0.195.
}

\input{figs/diversity2}
\paragraph{Bounding box distribution.}
We visualize \textit{bbox distributions} in Figure~\ref{fig:reb_diversity2} for object placement in Cityscapes images
and observe that our method identifies more diverse ROIs than TopNet. 
Our method sufficiently covers a wide range of reasonable locations, therefore cases of reasonable placements being missed are less likely to occur.

\paragraph{Decoder attention visualization.}
In \Cref{fig:attention}, we visualize the decoder attention of our trained model, revealing that the model attends to different regions for detecting objects of different categories. We observe that the decoder’s attention tends to focus on local extremities. The attention visualization aligns with our expectations: different queries detect different locations, such as vehicles being placed near the roadside or other vehicles, pedestrians being placed along the sidewalk, and bicycles either riding on the road or parked at the street side. 

%% file: tabs/recomposition.tex
\begin{table*}[t!]
\begin{center}
\scriptsize
\resizebox{\linewidth}{!}{
\begin{tabular}{l r r r r r r r r}
\toprule
& \multicolumn{4}{c}{Cityscapes} & \multicolumn{4}{c}{OPA}  \\
& IOU50@1 ($\uparrow$) & IOU@1 ($\uparrow$) & IOU50@5 ($\uparrow$) & IOU@5 ($\uparrow$) &  IOU50@1 ($\uparrow$) & IOU@1 ($\uparrow$) & IOU50@5 ($\uparrow$) & IOU@5 ($\uparrow$)\\
\midrule
PlaceNet (ECCV'20)~\cite{zhang2020learning} & 0 & 0.045 & 0 & 0.045 & 2.76 & 0.116 & 10.09 & 0.225 \\
GracoNet (ECCV'22)~\cite{SiyuanZhou2022LearningOP} & --- & --- & --- & --- &  2.49 & 0.131 & 16.60 & 0.248 \\
SAC-GAN (IEEE TVCG'22)~\cite{zhou2022sac} & 0.806 & 0.082 & 1.08 & 0.085 & --- & --- & --- & --- \\
TopNet (CVPR'23)~\cite{zhu2023topnet} & 0.807 & 0.045 & 1.61 & 0.070 & 11.55 & {\bf 0.197} & 15.95 & 0.241 \\
\ApproachName (ours) & {\bf 1.74} & {\bf 0.197} & {\bf 6.09} & {\bf 0.281} & {\bf 11.60} & {\bf 0.197} & {\bf 22.41} & {\bf 0.281} \\
\bottomrule
\end{tabular}
}
\end{center}
\vspace{-5mm}
\caption{
\textbf{Quantitative results of object reposition} on Cityscapes and OPA datasets, evaluated by IOU50 (\%), top-1 and top-5 IOU. 
}
\label{table:recomposition}
\end{table*}

%% file: figs/recomposition.tex
\begin{figure*}[t!]
\begin{center}
\includegraphics[width=1.0\linewidth]{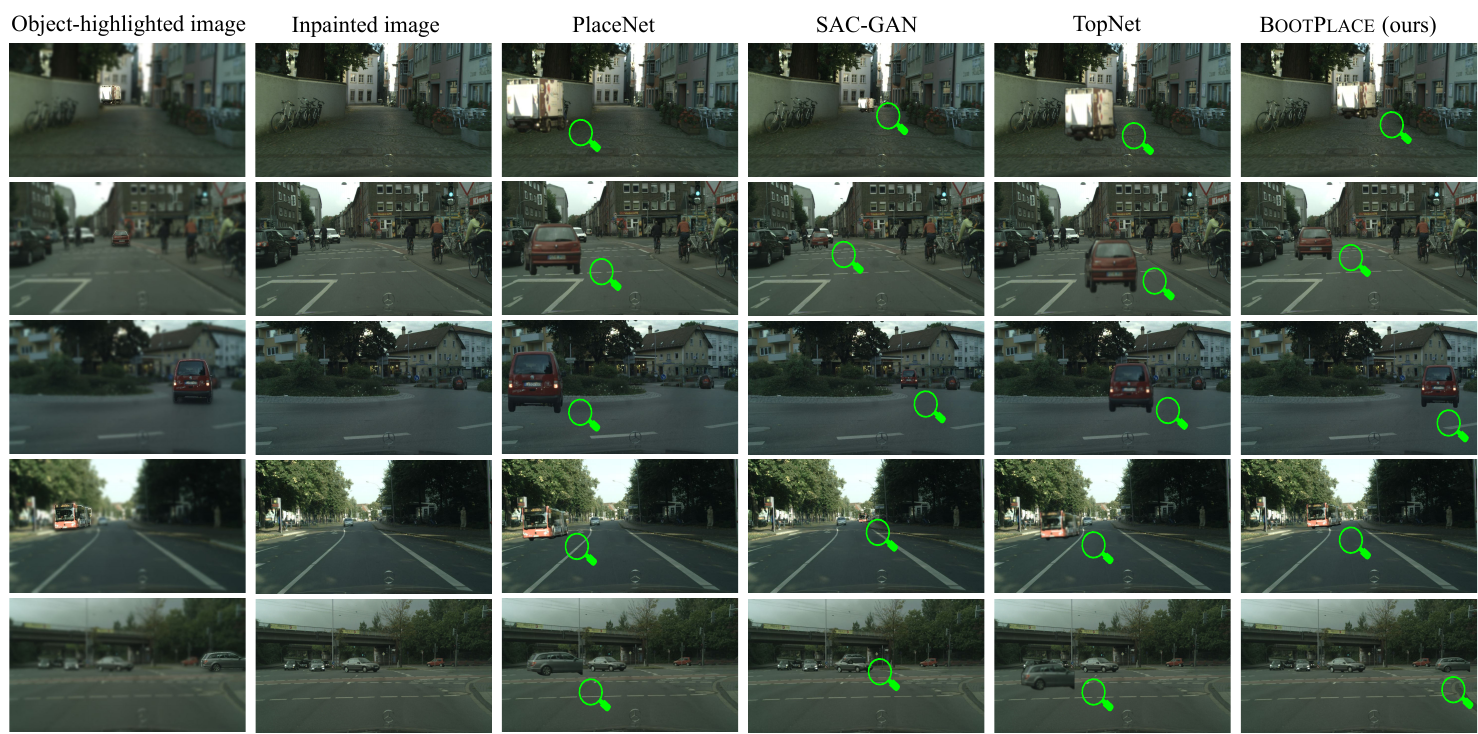}
\end{center}
\vspace{-5mm}
\caption{
\textbf{Qualitative results of object reposition} on Cityscapes dataset. Zoom in to see visual details. 1st column: original images with highlighted compositing objects; 2nd column: inpainted images after object subtraction. 
}
\label{fig:recomposition}
\end{figure*}

%% file: figs/recomposition_OPA.tex
\begin{figure*}[t!]
\begin{center}
\includegraphics[width=1.0\linewidth]{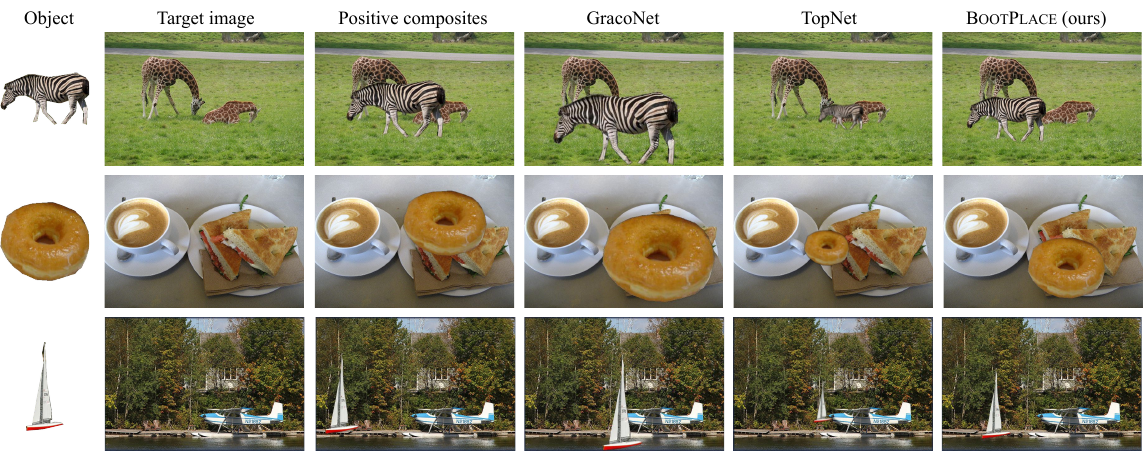}
\end{center}
\vspace{-5mm}
\caption{
\textbf{Qualitative results of object reposition}
on OPA dataset. ``Positive composites'' are annotated good-quality composites from OPA. SAC-GAN is excluded as it requires semantic maps for training.
}
\label{fig:recomposition_OPA}
\end{figure*}

%% file: figs/composition.tex
\begin{figure*}[t!]
\begin{center}
\includegraphics[width=1.0\linewidth]{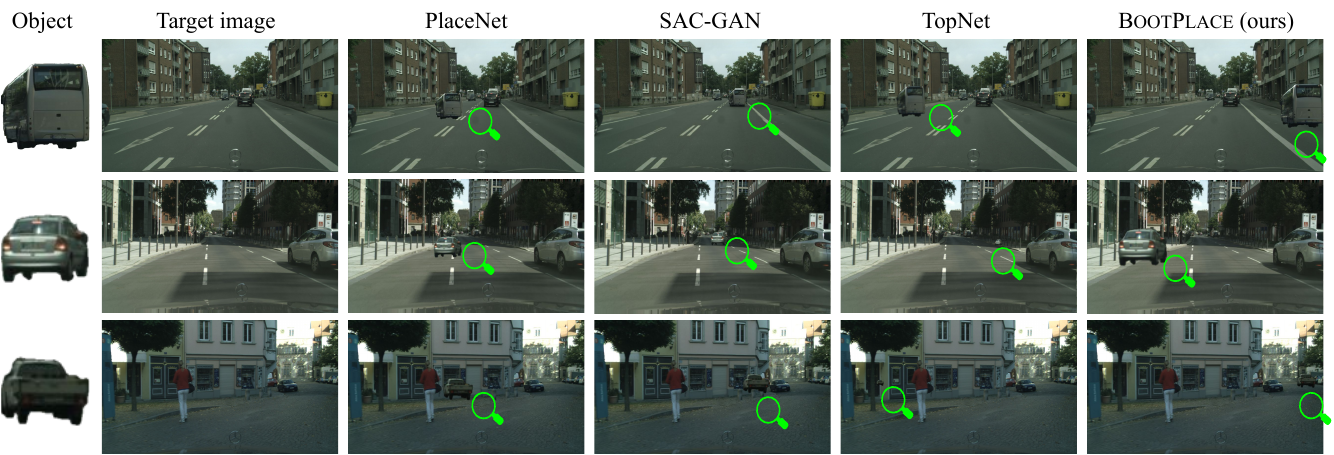}
\end{center}
\vspace{-4mm}
\caption{
\textbf{Qualitative results of object placement} on Cityscapes dataset. Objects are randomly chosen from Cityscapes testing set. 
}
\label{fig:composition}
\end{figure*}

%% file: figs/generalizability.tex
\begin{figure*}[t!]
\begin{center}
\includegraphics[width=1.0\linewidth]{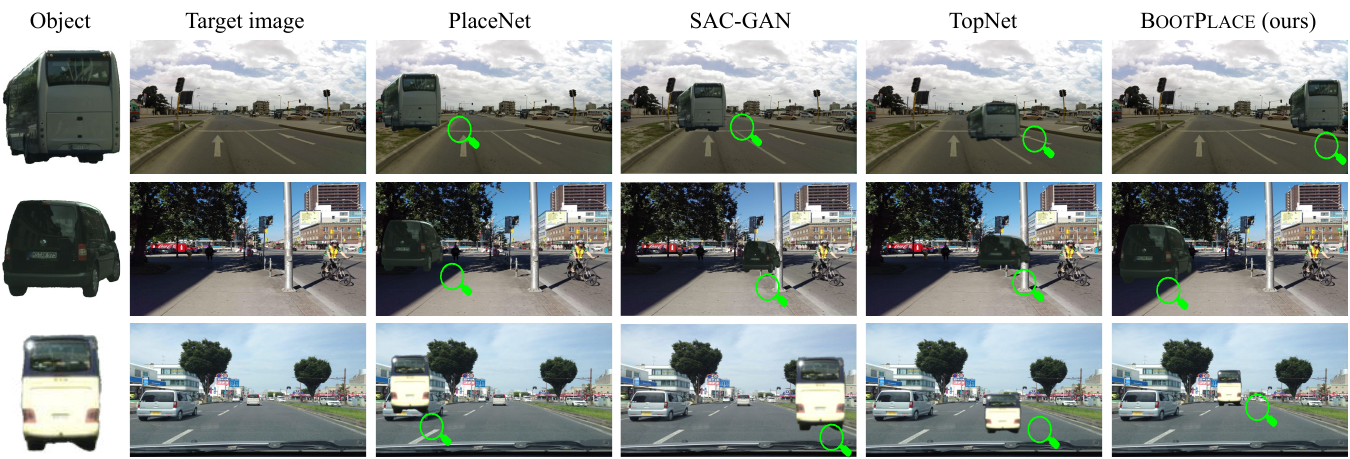}
\end{center}
\vspace{-4mm}
\caption{
\textbf{Qualitative results of object placement} on Mapillary Vistas dataset. Objects are randomly chosen from Cityscapes testing set.
}
\label{fig:generalizability}
\end{figure*}

%% file: tabs/composition.tex
\begin{table}[t!]
\begin{center}
\resizebox{\linewidth}{!}{
\begin{tabular}{l  r r r r r}
\toprule
&  \multicolumn{2}{c}{Plausibility} & \multicolumn{3}{c}{Diversity}\\
& \multicolumn{2}{c}{User study ($\uparrow$)} & \multicolumn{3}{c}{Mean$\pm$std ($\uparrow$)}\\
& Cityscapes & MV & scale & x-center & y-center \\
\midrule
PlaceNet~\cite{zhang2020learning} & 0.183 & 0.133 & 0.204$\pm$0.0008 & 0.287$\pm$0.0018 & 0.465$\pm$0.0005\\
SAC-GAN~\cite{zhou2022sac} & 0.269 & 0.285 & 0.156$\pm$0.0092 & 0.488$\pm$0.0405& 0.436$\pm$0.0056\\
TopNet~\cite{zhu2023topnet} & 0.246 & 0.260 & 0.079$\pm$0.0131&0.372$\pm$0.0818 & 0.464$\pm$0.0138\\
\ApproachName (ours) & \textbf{0.303} & \textbf{0.323} & 0.310$\pm$\textbf{0.1235} & 0.255$\pm$\textbf{0.1082} & 0.495$\pm$\textbf{0.0328}\\
\bottomrule
\end{tabular}
}
\end{center}
\vspace{-5mm}
\caption{
\textbf{Quantitative comparisons of car placement} on Cityscapes and Mapillary Vistas (MV) datasets.
}
\label{table:composition}
\end{table}

%% file: figs/attention.tex
\begin{figure*}[t!]
\begin{center}
\includegraphics[width=1\linewidth]{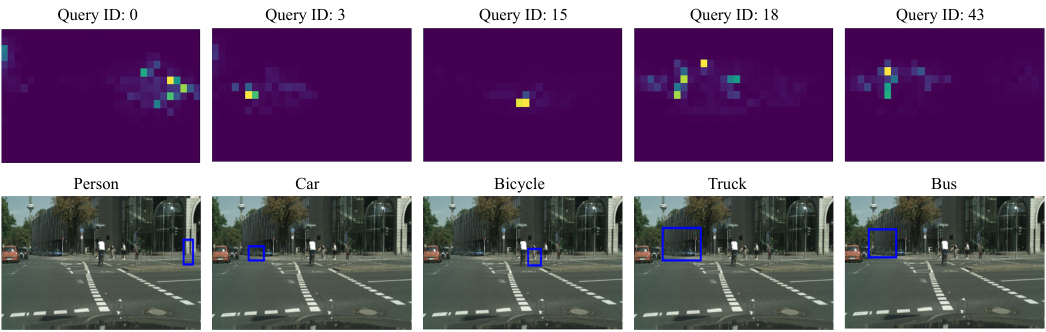}
\end{center}
\vspace{-4mm}
\caption{
\textbf{Visualizing decoder attention}
for every detected region for object placement (images from Cityscapes val set), which part of the image the model was looking at to predict this specific bounding box and class. Best viewed in color.
}
\vspace{-2mm}
\label{fig:attention}
\end{figure*}

%% file: figs/overfit.tex
\setlength{\intextsep}{0pt}
\begin{wrapfigure}{r}{0.24\textwidth}
    \centering
    \hspace{-6mm}
        \resizebox{0.26\textwidth}{!}{%
			\includegraphics[width=1\columnwidth]{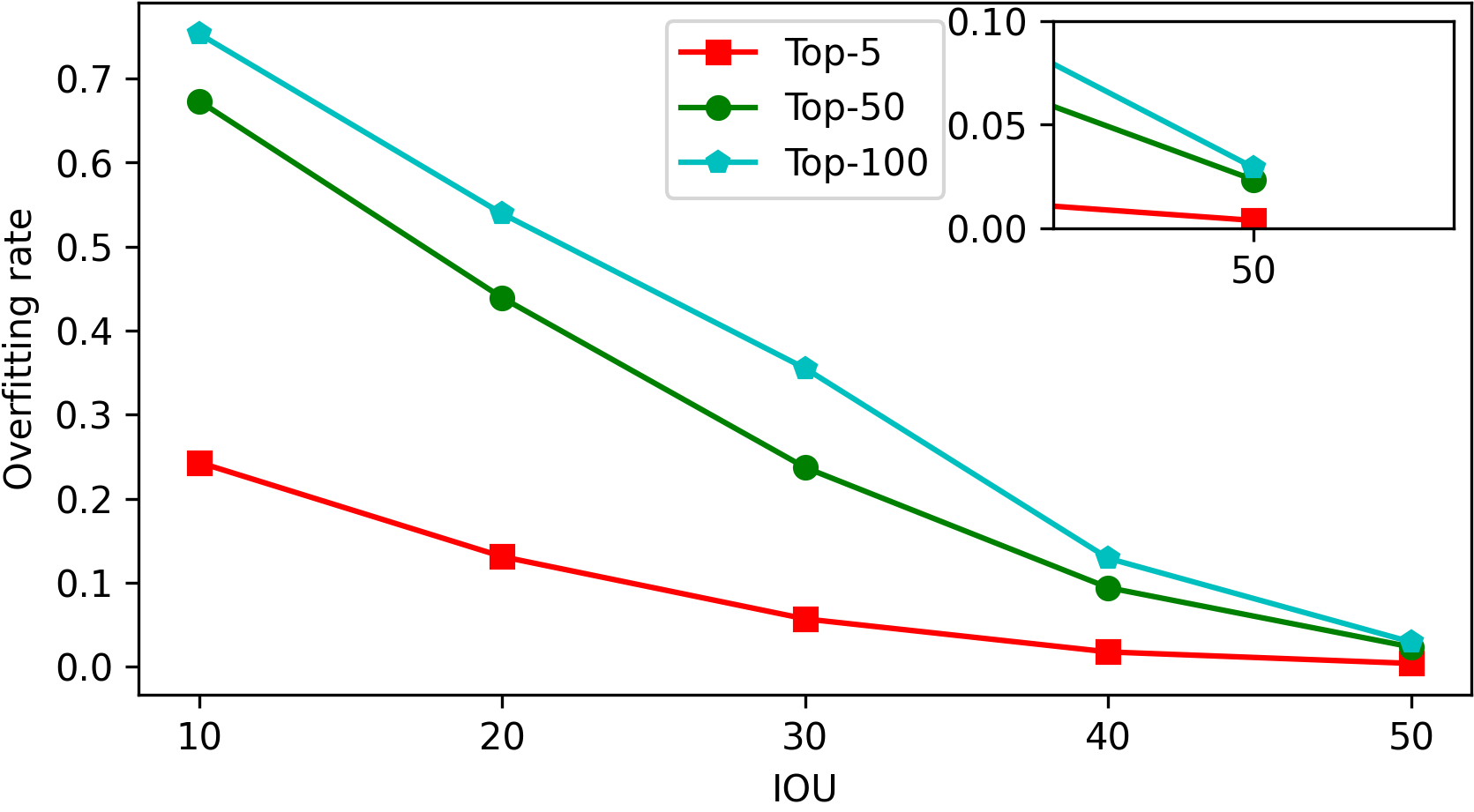}
		}
        \vspace{-2.5mm}
  \caption{Overfitting rate vs IOU.}
  \label{fig:reb_overfit}
\end{wrapfigure}

%% file: tabs/ablation.tex
\begin{table}[t!]
\begin{center}
\vspace{-2mm}
\resizebox{\linewidth}{!}{
\begin{tabular}{l r r r r r r}
\toprule
& W/o smooth. & W/o augment. &Pos. contrast.  & Single label & W/o LE & Full model\\
\midrule
IOU50@1 ($\uparrow$) & 0 & 0.94 & 1.88 & 0 & 0 & {\bf 1.74}\\
IOU@1 ($\uparrow$) & 0 & 0.056 & 0.049 & 0.167 & 0.069 & {\bf 0.197}\\
IOU50@5 ($\uparrow$) & 4.23 & 3.29 & 4.69 & 2.61 & 0 & {\bf 6.09}\\
IOU@5 ($\uparrow$) & 0.082 & 0.121 & 0.125 & 0.239 & 0.086 & {\bf 0.281}\\
\bottomrule
\end{tabular}
}
\end{center}
\vspace{-5mm}
\caption{
\textbf{Ablation study} of object reposition on Cityscapes~\cite{cordts2016cityscapes} dataset, evaluated by IOU50 (\%), top-1 and top-5 IOU. \hang{LE stands for location encoder. }
}
\vspace{-5mm}
\label{table:ablation}
\end{table}

%% file: figs/diversity2.tex
\begin{figure}[t!]
\begin{center}
\includegraphics[width=0.49\linewidth]{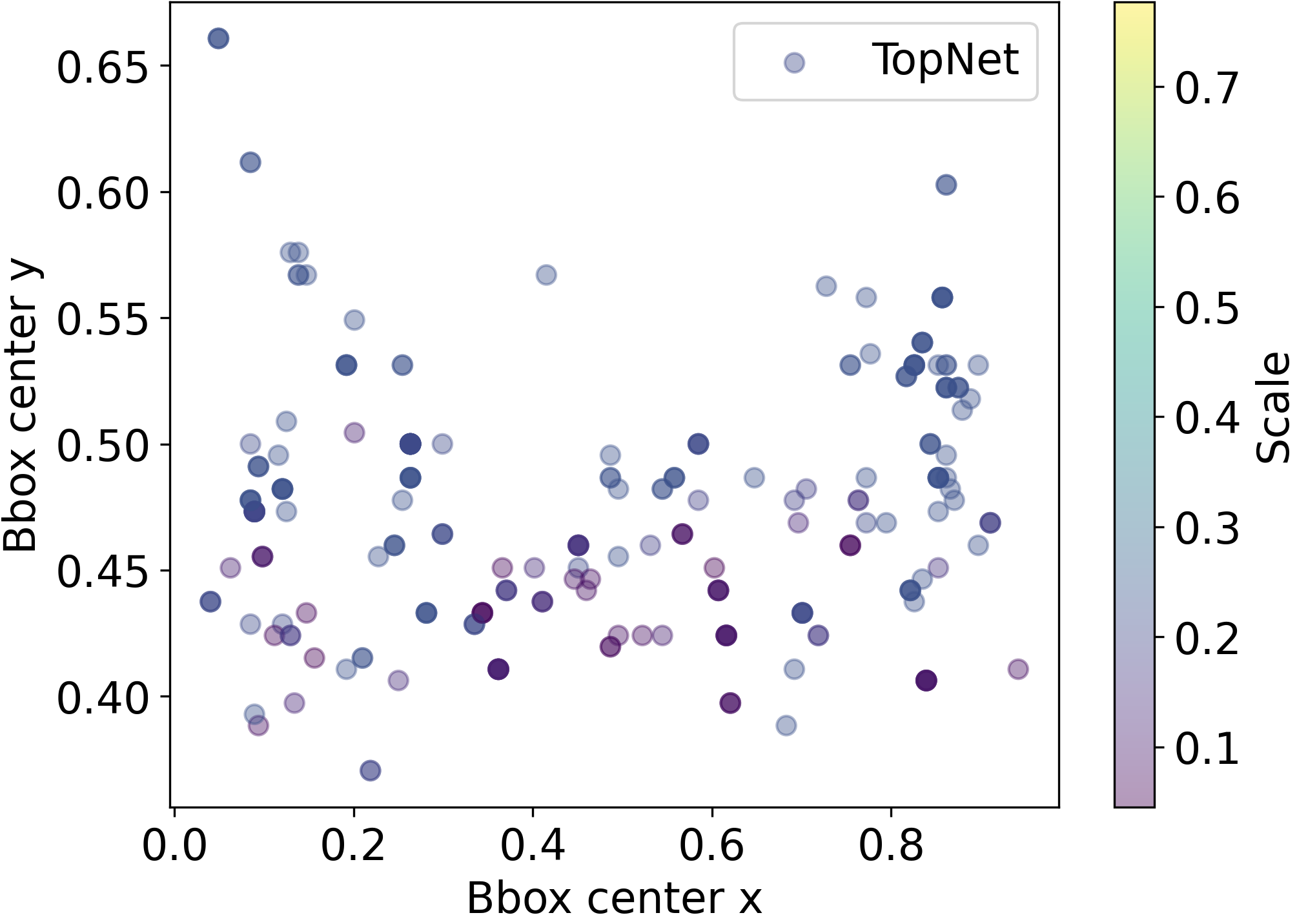}
\includegraphics[width=0.49\linewidth]{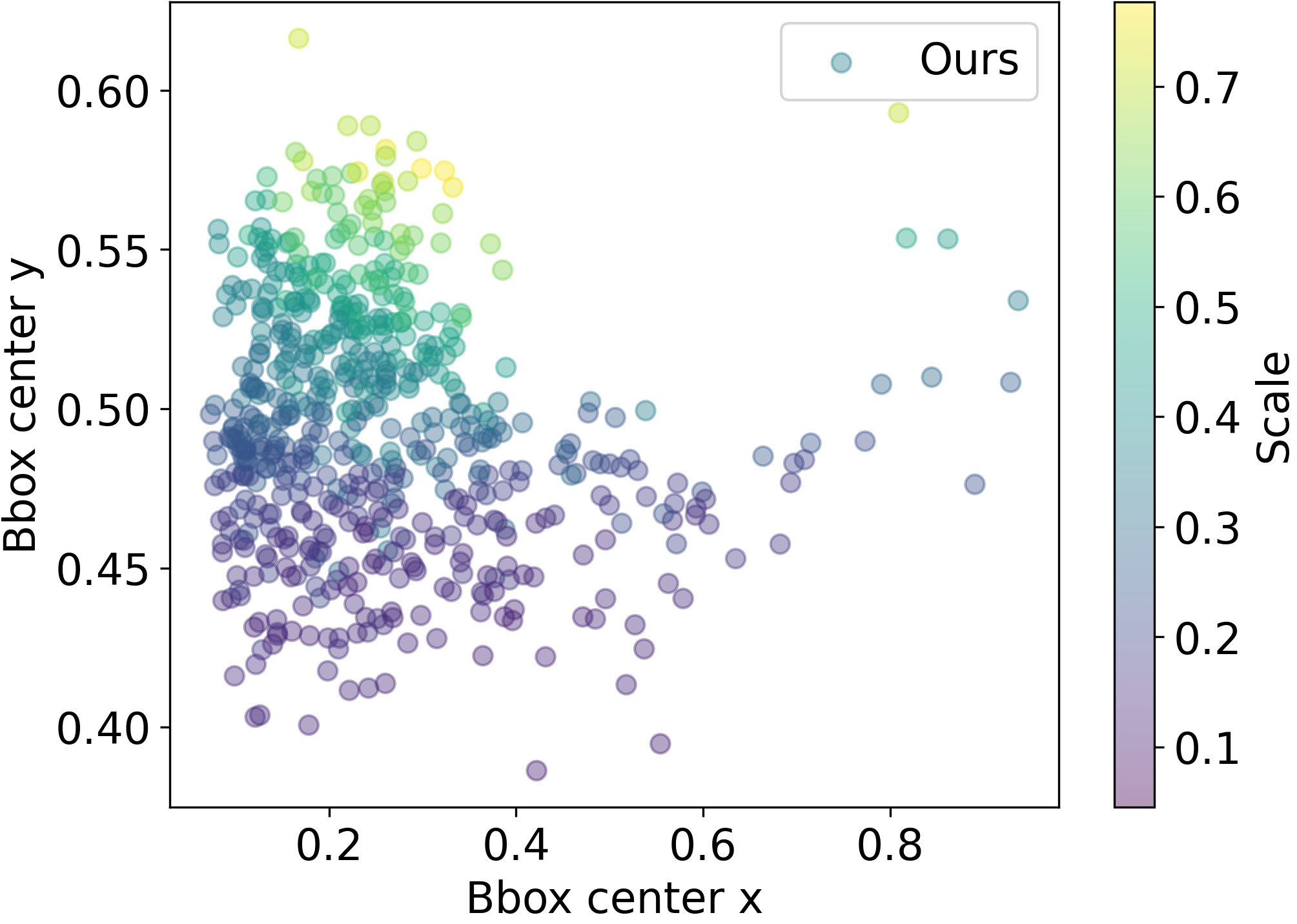}
\end{center}
\vspace{-5mm}
\caption{Bbox distribution. Color intensity reflects bbox scales. 
    }
    \label{fig:reb_diversity2}
\end{figure} 

%% file: sec/5_conclusions.tex
\section{Discussions, limitations and future work}
\label{sec:discuss}

\input{figs/failure.tex}

We have introduced \ApproachName, a placement-by-detection paradigm based on detection transformers and association networks. 
By leveraging multi-object supervision to detect regions of interest, our approach significantly enhances object placement performance. 
The proposed association network enables the differentiable association between compositing objects and the detected regions. 
Additionally, we introduce a bootstrapped training strategy that increases data diversity.

Similar to most object detection methods, our approach detects locations \textit{in parallel}, limiting its suitability for sequential object placement. 
As shown in \Cref{fig:failure}, this leads to undesirable occlusions, such as a car colliding with a curb (left) and overlapping cars (right). 
Additionally, most object placement methods lack out-of-plane rotation and perspective transformation modeling, resulting in less realistic compositions. 
To overcome these challenges, we intend to develop autoregressive models to better handle the sequential object placement problem.

%% file: figs/failure.tex
\begin{figure}[t!]
\begin{center}
\includegraphics[width=1.0\linewidth]{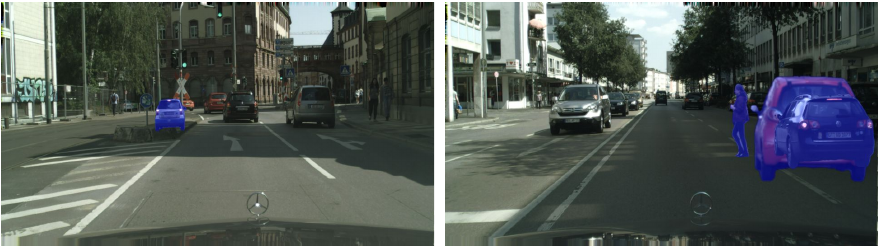}
\end{center}
\vspace{-5mm}
\caption{
\textbf{Failure cases.}
Composed objects are rendered in blue.
}
\vspace{-3mm}
\label{fig:failure}
\end{figure}

%% file: sec/6_acks.tex
% \noindent\textbf{Acknowledgments:} This work was partly supported by NSERC Discovery, CFI-JELF, NSERC Alliance, Alberta Innovates and PrairiesCan grants. Part of this work was conducted at the Simon Fraser University and we thank Prof. Hao Zhang for insightful discussions. 

\section*{Acknowledgments}
This work was partly supported by NSERC Discovery, CFI-JELF, NSERC Alliance, Alberta Innovates and PrairiesCan grants. Part of this work was conducted at the Simon Fraser University and we thank Prof. Hao Zhang for insightful discussions. 
\par

%% file: sec/7_supp.tex
\clearpage
\setcounter{page}{1}
\maketitlesupplementary

\section{Network and parameter details}
\label{section:supp_details}

For the image encoder, We use ImageNet-pretrained ResNet-50 with frozen batchnorm layers and discard the last classification layer as the CNN backbone. 
The Transformer encoder contains 6 blocks and the Transformer decoder contains 6 block. Each attention layer has 8 attention heads. Additive dropout of 0.1 is applied after every multi-head attention and FFN before layer normalization. The weights are randomly initialized with Xavier initialization. The intermediate size of the feedforward layers in the transformer blocks is set 2048 and the size of the embeddings $d$ in the transformer is set 256. 
The number of object (region) queries $N$ is set to 100 and the maximum number of objects $M$ is set to 120. 
For the bounding box encoder, we utilize 2-layer MLP which transforms the bounding box embedding into 256 dimensiton and multiple it with the embedding obtained from the Transformer decoder, and then use 3-layer MLP to map it to 4-dim embeddings.

\section{Computational cost}
To compare the computation costs of different methods, we show the number of Params, FLOPs and inference time of PlaceNet, SAC-GAN, TopNet and \ApproachName in Table~\ref{table:cost}. 
Our method requires slightly more parameters than the other methods. The theoretical computation cost (FLOPs) of our method is 7x larger, but is similar to DETR-based detection models. We also tested the inference time on the Cityscapes validation set using an Nvidia GeForce GTX TITAN X with a batch size of 1. The inference time per sample of our method is less than 1 second, significantly faster than TopNet (2.7 seconds) and similar to PlaceNet. Therefore, despite the higher FLOPs, the computational complexity of our method is manageable and affordable, with efficient real-world inference times.

\input{figs/multi_composition.tex}
\input{tabs/computation_cost.tex}
\input{figs_supp/loss.tex}

\input{figs_supp/data.tex}

\input{figs_supp/harmonization_data.tex}
\input{figs_supp/harmonization.tex}

\section{Loss function analysis}
In \Cref{fig:loss}, we compare the impact of various loss functions. 
The regression loss, with sparse annotation, poses challenges in training the model effectively.
Gaussian assigned loss overlooks the impact of scaling and fails to accommodate multi-peak distributions for possible placements.
Sparse contrastive loss supports the fluctuation of neighboring placements but lacks accurate constraints for complex scenes with location-varying placements.
Our proposed loss function is derived from bounding box loss using Generalized IOU loss, offering more precise constraints on box scaling.

\section{Multi-object placement}
Compared to single-object placement, multi-object placement is significantly more challenging as it necessitates an understanding of the prior state of composed objects, scene objects and background image. 
Though our network makes parallel bounding box predictions, it has learned a robust correspondence between objects and their associated regions. 
In \Cref{fig:multi_composition}, we illustrate the potential for composing three objects into street scenes, showing the capability of our network to learn object orientation and the distribution of various object categories.

\section{Dataset construction}
\label{section:data_prepare}
In Figure~\ref{fig:data}, we illustrate the data construction process for the Cityscapes~\cite{cordts2016cityscapes} dataset, which is applicable to other datasets. 
We start with the source image~\circled{1} and employ a pretrained MaskFormer~\cite{cheng2021per} model for panoptic semantic segmentation~\circled{2}, jointly performing semantic and instance segmentation. 
Scene primitives are manually categorized into object classes, including \texttt{car, person, rider, train, bus, bicycle, truck, motorcycle}, resulting in binary object masks~\circled{3}.
To obtain object-free backgrounds, we dilate the binary object masks to address boundary inaccuracies and obtain dilated object masks~\circled{4}. 
The next step involves using pretrained LaMa~\cite{suvorov2022resolution} inpainting model to remove objects, yielding inpainted images~\circled{7}. 
As many segmentation models tend to classify shadows as background, we manually remove these shadows using an online PhotoKit tool, refining the background image to obtain corrected inpainted images~\circled{8}.
Simultaneously, we create an object pool~\circled{5} consisting of both intact objects and those that are partially occluded, each with varying resolutions. 
After manual curation, we retain only the intact objects, resulting in an intact object pool~\circled{6} with their bounding box coordinates. 
After data cleaning, we construct a multi-object dataset including 2,953 training images with 22,270 objects and their corresponding ground-truth labels, as well as 372 testing images with 2,713 objects.

\input{tabs/comparison_topnet.tex}
\input{figs_supp/postprocess}

\vspace{-0.5mm}
\hang{
\section{Method comparison with TopNet}
We provide comparisons between TopNet and \ApproachName in architecture and training strategy in Table~\ref{table:reb_comp_topnet}.
}

\vspace{-1mm}
\section{Image blending}
To address boundary artifacts arising from copy-paste object composition, we \hang{employ two distinct methods: (1) use} a diffusion model initially designed for image inpainting to harmonize boundaries. 
We finetune the Stable Diffusion Inpainting model without prompt conditioning~\footnote{https://github.com/lorenzo-stacchio/Stable-Diffusion-Inpaint/tree/1b44f2f9e4f233f68d48c56b68b9c111c1538d4d} on Cityscapes objects. 
This involves extracting object-centric patches and dilating their masks to create boundary masks, as depicted in \Cref{fig:harmonization_data}. 
Once a sufficient number of such patch-mask pairs are collected, we finetune the model to harmonize boundary region.
\Cref{fig:harmonization} shows the visual performance of image harmonization on Cityscapes samples. 
\hang{(2) combine placement learning with identity-preserving compositing methods such as  ObjectStitch~\cite{song2023objectstitch} for visual refinement, as shown in Figure~\ref{fig:reb_postprocess}.
}
This process significantly reduces boundary artifacts while naturally generating shadows, resulting in an enhanced level of realism compared to compositions without harmonization.

\vspace{-1mm}
\hang{
\section{Evaluation on FID and LPIPS}
\input{tabs/fid_lpips.tex}

In Table~\ref{table:fid_lpips}, we evaluate plausibility of composite images using FID and LPIPS on the Cityscapes dataset. 
We observe that both metrics are strongly correlated with bbox scale, where smaller bounding boxes result in fewer modifications to the image and correspondingly lower FID and FPIPS values. 
Therefore, they are not suitable metrics for evaluating placement quality, which are excluded from evaluation. 
}

\vspace{-1mm}
\section{More qualitative results}
We show qualitative results of object placement on Cityscapes dataset in \Cref{fig:comp_vis,fig:comp_vis2}, object reposition on Cityscapes dataset in \Cref{fig:recomp_vis,fig:recomp_vis2}, and object reposition on OPA dataset in \Cref{fig:recomp_vis_opa}.

\vspace{-1mm}
\section{More decoder attention visualization}
\input{figs_supp/composition_vis.tex}
\input{figs_supp/recomposition_vis.tex}
\input{figs_supp/recomposition_vis_opa.tex}

\input{figs_supp/decoder_vis.tex}

We provide additional visualization results showing the distribution of the detection decoder in \Cref{fig:decoder_vis1}.

%% file: figs/multi_composition.tex
\begin{figure}[th!]
\begin{center}
\includegraphics[width=1.0\linewidth]{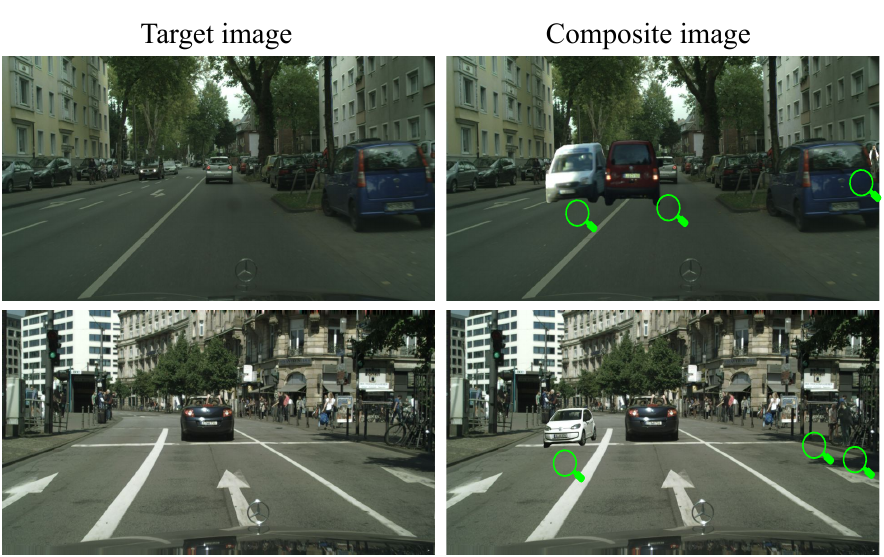}
\end{center}
\vspace{-2.5mm}
\caption{
\textbf{Multi-object placement.}
Two cars and one person (top), one car and two people (below) are composed into the scene. 
}
\label{fig:multi_composition}
\end{figure}

%% file: tabs/computation_cost.tex
\begin{table*}
\begin{center}
\scriptsize
\begin{tabular}{l r r r r}
\toprule
& PlaceNet (ECCV'20)~\cite{zhang2020learning} & SAC-GAN (IEEE TVCG'22)~\cite{zhou2022sac} & TopNet (CVPR'23)~\cite{zhu2023topnet} & \ApproachName (ours)\\
\midrule
\# Params (M) & 35.9 &  35.9 &  25.0 & 41.4\\
\# FLOPs (G) & 4.40 & 6.96 & 6.79 & 44.4 \\
Inference time (sec) & 0.68 & 0.1 & 2.7 & 0.27 \\
\bottomrule
\end{tabular}
\end{center}
\vspace{-2.5mm}
\caption{
Comparison of computational cost and model parameters tested on Cityscapes dataset. 
}
\label{table:cost}
\end{table*}

%% file: figs_supp/loss.tex
\begin{figure*}[t!]
\begin{center}
\includegraphics[width=1.0\linewidth]{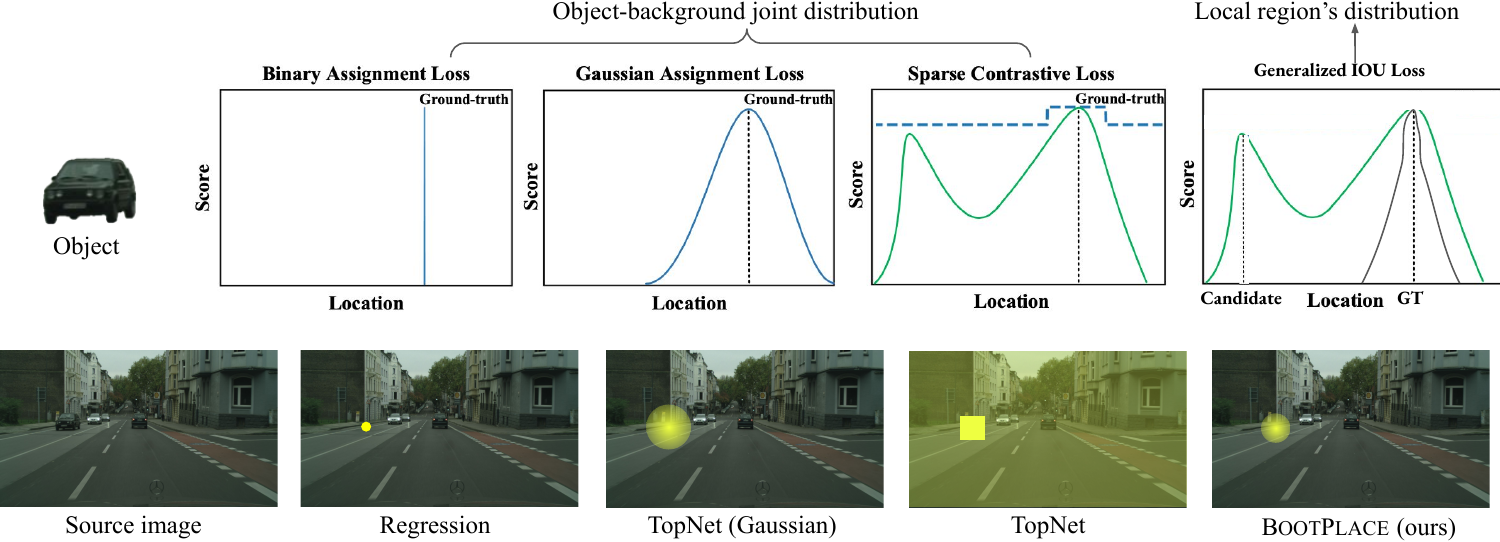}
\end{center}
\vspace{-2.5mm}
\caption{
\textbf{Different losses} exemplified in 1D space. The yellow marks, depicted with varying intensities, represent the constraint intensity. 
}
\label{fig:loss}
\end{figure*}

%% file: figs_supp/data.tex
\begin{figure*}[!htb]
\begin{center}
\includegraphics[width=1\linewidth]{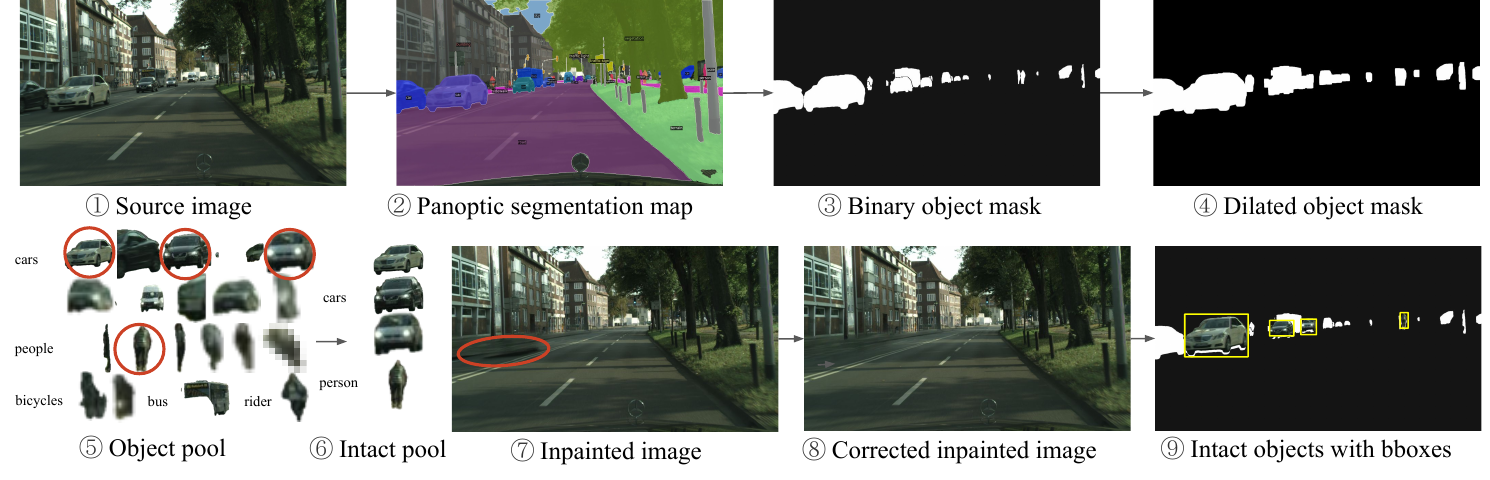}
\end{center}
\vspace{-2.5mm}
\caption{
\textbf{Data preparation}
of background inpainted images and corresponding scene objects processed from source images. 
}
\label{fig:data}
\end{figure*}

%% file: figs_supp/harmonization_data.tex
\begin{figure}[t!]
\begin{center}
\includegraphics[width=1\linewidth]{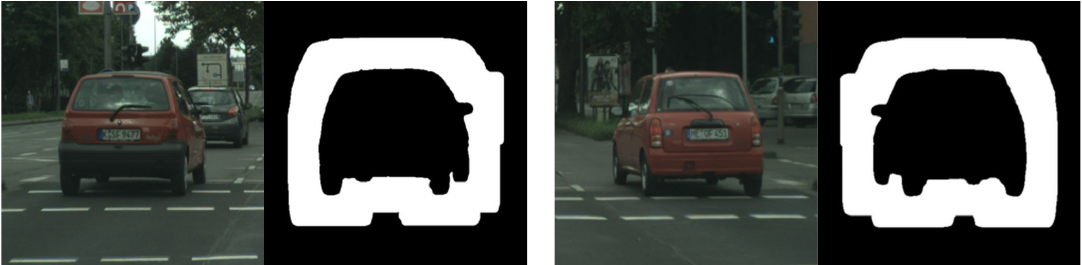}
\end{center}
\caption{
\textbf{Data preparation}
for boundary harmonization. The boundary dilated regions are automatically segmented by dilation of object silhouette. 
}
\label{fig:harmonization_data}
\end{figure}

%% file: figs_supp/harmonization.tex
\begin{figure*}[!htb]
\begin{center}
\includegraphics[width=1.0\linewidth]{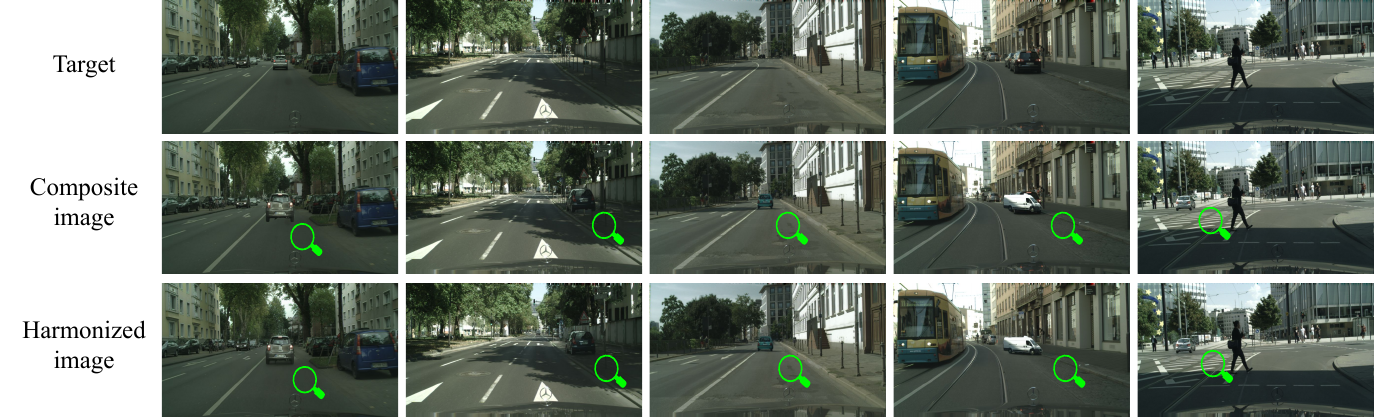}
\end{center}
\vspace{-4mm}
\caption{
\textbf{Harmonization results}
of composite Cityscapes samples. Zoom in to see visual details.
}
\label{fig:harmonization}
\end{figure*}

%% file: tabs/comparison_topnet.tex
\begin{table}[t!]
\begin{center}
\resizebox{\linewidth}{!}{
% \tiny
\begin{tabular}{l r r}
\toprule
& TopNet & \ApproachName (Ours) \\
\midrule
Encoder & ViT-small & CNN 
 + MLP + Transformer encoders \\
Decoder & 2D upsampling&Transformer decoder + MLPs \\
Output & 3D heatmap & Bbox + class predictions\\
Losses & Sparse contrastive + range & Bbox regression + class prediction + association\\
Training & AdamW optimizer + lr=1e-5 & AdamW optimizer + le= 4e-4\\
\bottomrule
\end{tabular}
}
\end{center}
\vspace{-4mm}
\caption{
TopNet \textit{vs} \ApproachName (ours).
}
\label{table:reb_comp_topnet}
\end{table}

%% file: figs_supp/postprocess.tex
\begin{figure}[!htb]
  \centering
  \includegraphics[width=1.0\linewidth]{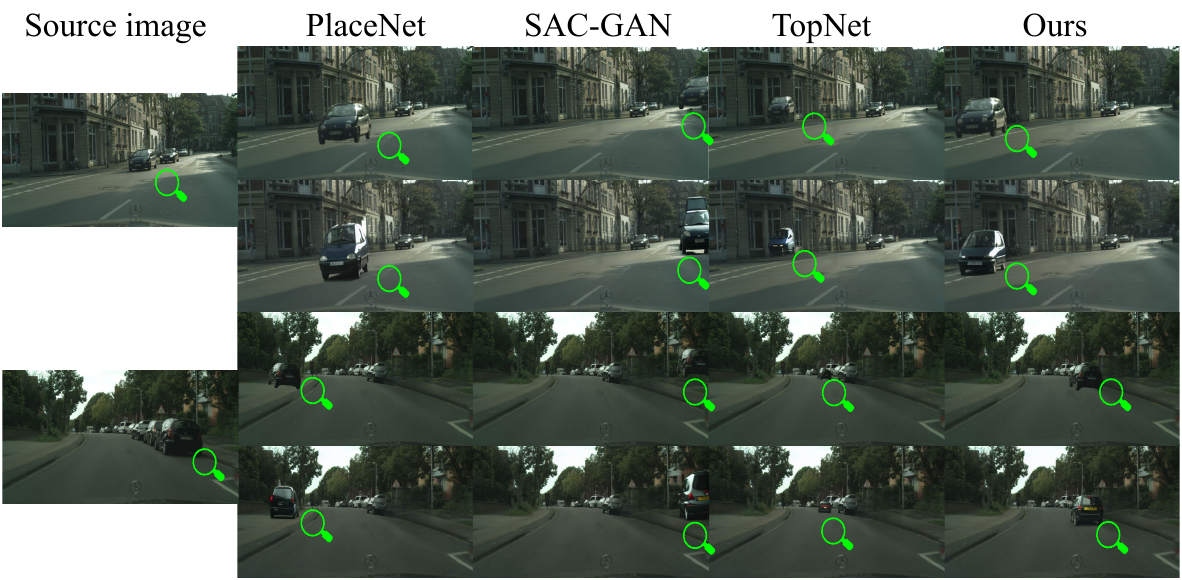}
  \vspace{-4mm}
   \caption{Comparison of w/o (row 1 and 3) \textit{vs} w/ (row 2 and 4) object compositing for two examples on object replacement task. }
   \label{fig:reb_postprocess}
\end{figure}

%% file: tabs/fid_lpips.tex
\begin{table}[t!]
\begin{center}
\vspace{-4mm}
\resizebox{\linewidth}{!}{
\begin{tabular}{l r r r r r}
\toprule
& \multicolumn{2}{c}{Copy-paste} & \multicolumn{2}{c}{ObjectStitch}\\
& FID ($\downarrow$) & LPIPS ($\downarrow$) & FID ($\downarrow$) & LPIPS ($\downarrow$) & \multicolumn{1}{c}{Scale}\\
\midrule
PlaceNet~\cite{zhang2020learning} & 52.02 & 0.088 & 77.67 & 0.217 & 0.204\\
SAC-GAN~\cite{zhou2022sac} & 42.89 & 0.066 & 63.44 & 0.194 & 0.156\\
TopNet~\cite{zhu2023topnet} & 38.21 & 0.043 & 49.74 & 0.189 & 0.079\\
\ApproachName (ours) & 58.74 & 0.105 & 79.50 & 0.246 & 0.310\\
\bottomrule
\end{tabular}
}
\end{center}
\vspace{-4mm}
\caption{
Quantitative comparison using FID and LPIPS.
}
\label{table:fid_lpips}
\end{table}

%% file: figs_supp/composition_vis.tex
\begin{figure*}[t!]
\begin{center}
\includegraphics[width=1.0\linewidth]{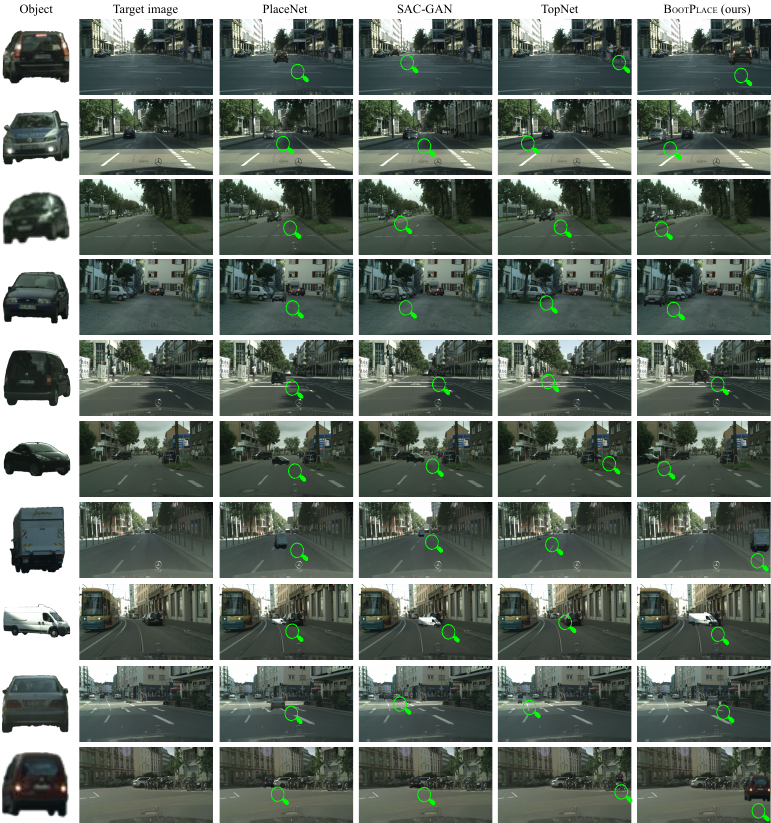}
\end{center}
\vspace{-4mm}
\caption{
\textbf{Qualitative results of single object placement} on Cityscapes dataset. Objects are randomly chosen from its testing set. 
}
\label{fig:comp_vis}
\end{figure*}

\begin{figure*}[t!]
\begin{center}
\includegraphics[width=1.0\linewidth]{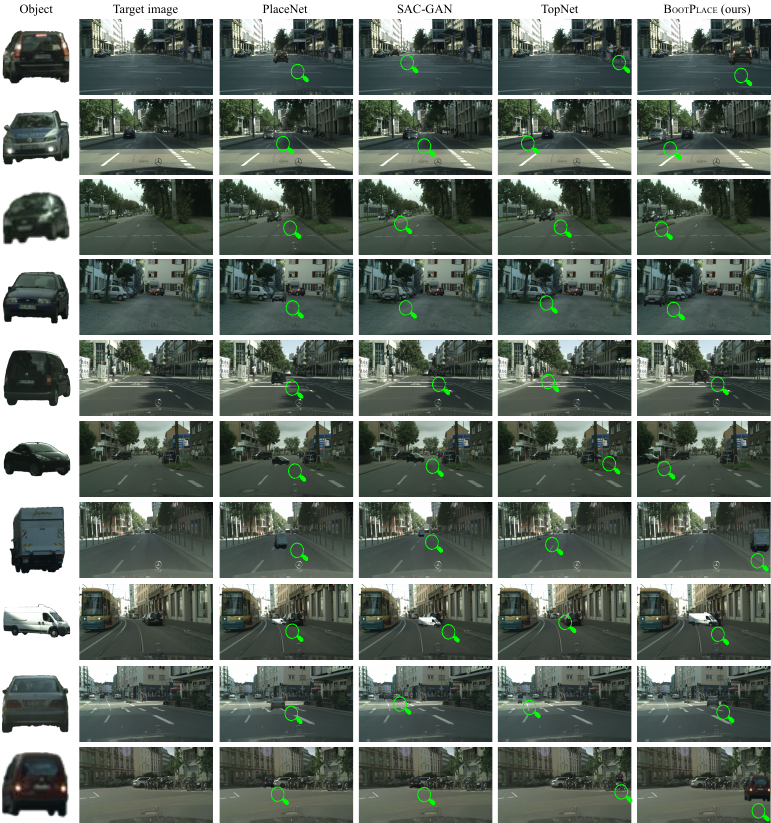}
\end{center}
\vspace{-4mm}
\caption{
\textbf{Qualitative results of object placement} on Cityscapes dataset. Objects are randomly chosen from its testing set. 
}
\label{fig:comp_vis2}
\end{figure*}

%% file: figs_supp/recomposition_vis.tex
\begin{figure*}[t!]
\begin{center}
\includegraphics[width=1.0\linewidth]{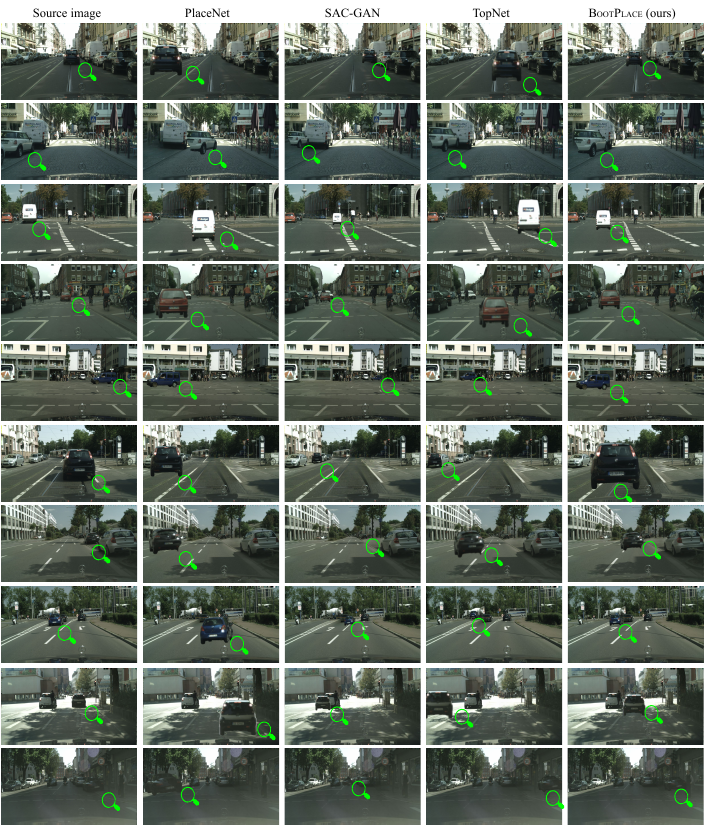}
\end{center}
\vspace{-4mm}
\caption{
\textbf{Qualitative results of object reposition} on Cityscapes dataset. 
}
\label{fig:recomp_vis}
\end{figure*}

\begin{figure*}[t!]
\begin{center}
\includegraphics[width=1.0\linewidth]{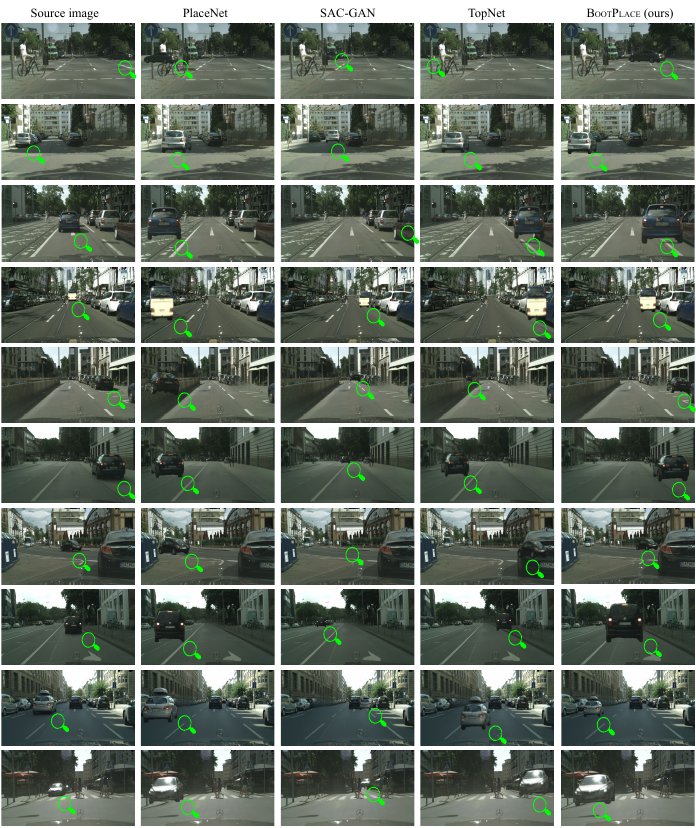}
\end{center}
\vspace{-4mm}
\caption{
\textbf{Qualitative results of object reposition} on Cityscapes dataset. 
}
\label{fig:recomp_vis2}
\end{figure*}

%% file: figs_supp/recomposition_vis_opa.tex
\begin{figure*}[t!]
\begin{center}
\includegraphics[width=1.0\linewidth]{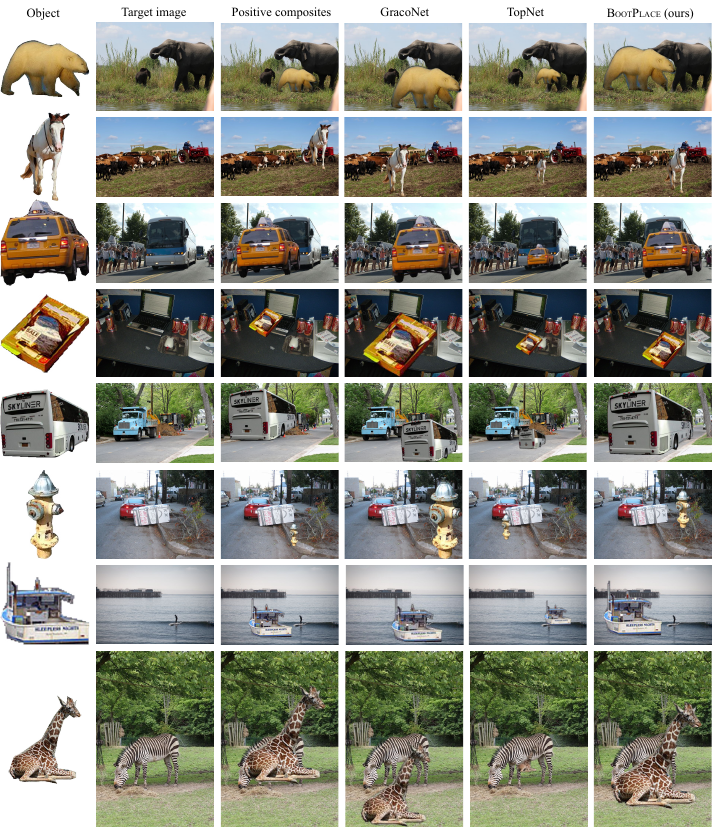}
\end{center}
\vspace{-4mm}
\caption{
\textbf{Qualitative results of object reposition} on OPA dataset.
}
\label{fig:recomp_vis_opa}
\end{figure*}

%% file: figs_supp/decoder_vis.tex
\begin{figure*}[t!]
\begin{center}
\includegraphics[width=0.98\linewidth]{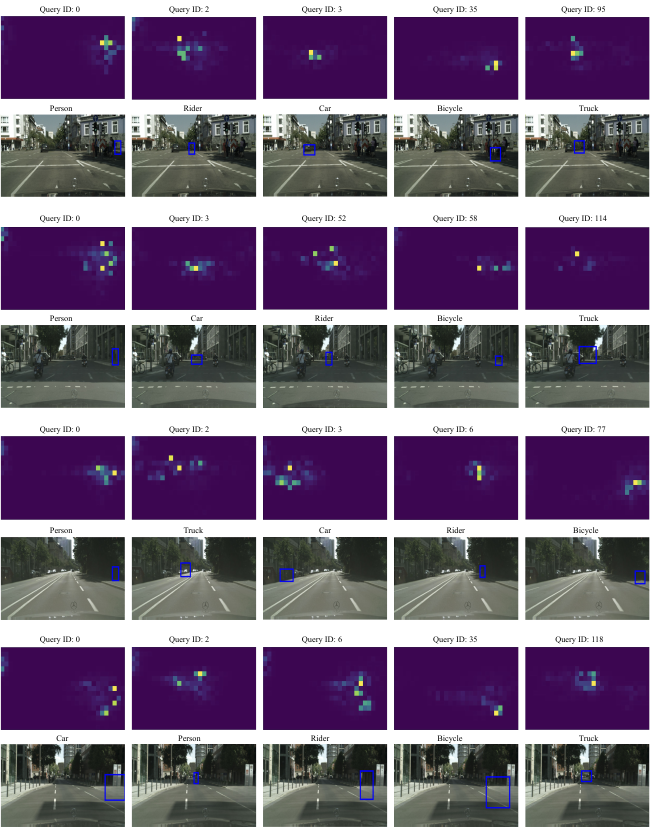}
\end{center}
\vspace{-4mm}
\caption{
\textbf{Decoder attention visualization}
of Cityscapes samples.
}
\label{fig:decoder_vis1}
\end{figure*}